\def\year{2021}\relax
\documentclass[letterpaper]{article} 
\usepackage{aaai21}  
\usepackage{times}  
\usepackage{helvet} 
\usepackage{courier}  
\usepackage[hyphens]{url}  
\usepackage{graphicx} 
\def\UrlFont{\rm}  
\usepackage{natbib}  
\usepackage{caption} 
\usepackage{amsmath}
\usepackage{multirow}      
\usepackage{multicol}
\usepackage{algorithm}
\usepackage{amssymb}
\usepackage[noend]{algpseudocode}
\usepackage[table]{xcolor}
\usepackage{bbm}
\usepackage{pbox}
\usepackage{comment}
\usepackage[switch]{lineno} 
\title{Role of Orthogonality Constraints in \\Improving Properties of Deep Networks for Image Classification}
\author{Hongjun Choi, Anirudh Som, Pavan Turaga}
\affiliations{\\Geometric Media Lab\\Arizona State University}

\begin{document}

\maketitle

\begin{abstract}
Standard deep learning models that employ the categorical cross-entropy loss are known to perform well at image classification tasks. However, many standard models thus obtained often exhibit issues like feature redundancy, low interpretability, and poor calibration. A body of recent work has emerged that has tried addressing some of these challenges by proposing the use of new regularization functions in addition to the cross-entropy loss. In this paper, we present some surprising findings that emerge from exploring the role of simple orthogonality constraints as a means of imposing physics-motivated constraints common in imaging. We propose an Orthogonal Sphere (\emph{OS}) regularizer that emerges from physics-based latent-representations under simplifying assumptions. Under further simplifying assumptions, the \emph{OS} constraint can be written in closed-form as a simple orthonormality term and be used along with the cross-entropy loss function. The findings indicate that orthonormality loss function results in a) rich and diverse feature representations, b) robustness to feature sub-selection, c)  better semantic localization in the class activation maps, and d) reduction in model calibration error. We demonstrate the effectiveness of the proposed \emph{OS} regularization by providing quantitative and qualitative results on four benchmark datasets - CIFAR10, CIFAR100, SVHN and tiny ImageNet.
\end{abstract}

\section{Introduction}\label{sec:introduction}

\begin{figure}[t!]
	\begin{center}
		\includegraphics[width=.99\linewidth]{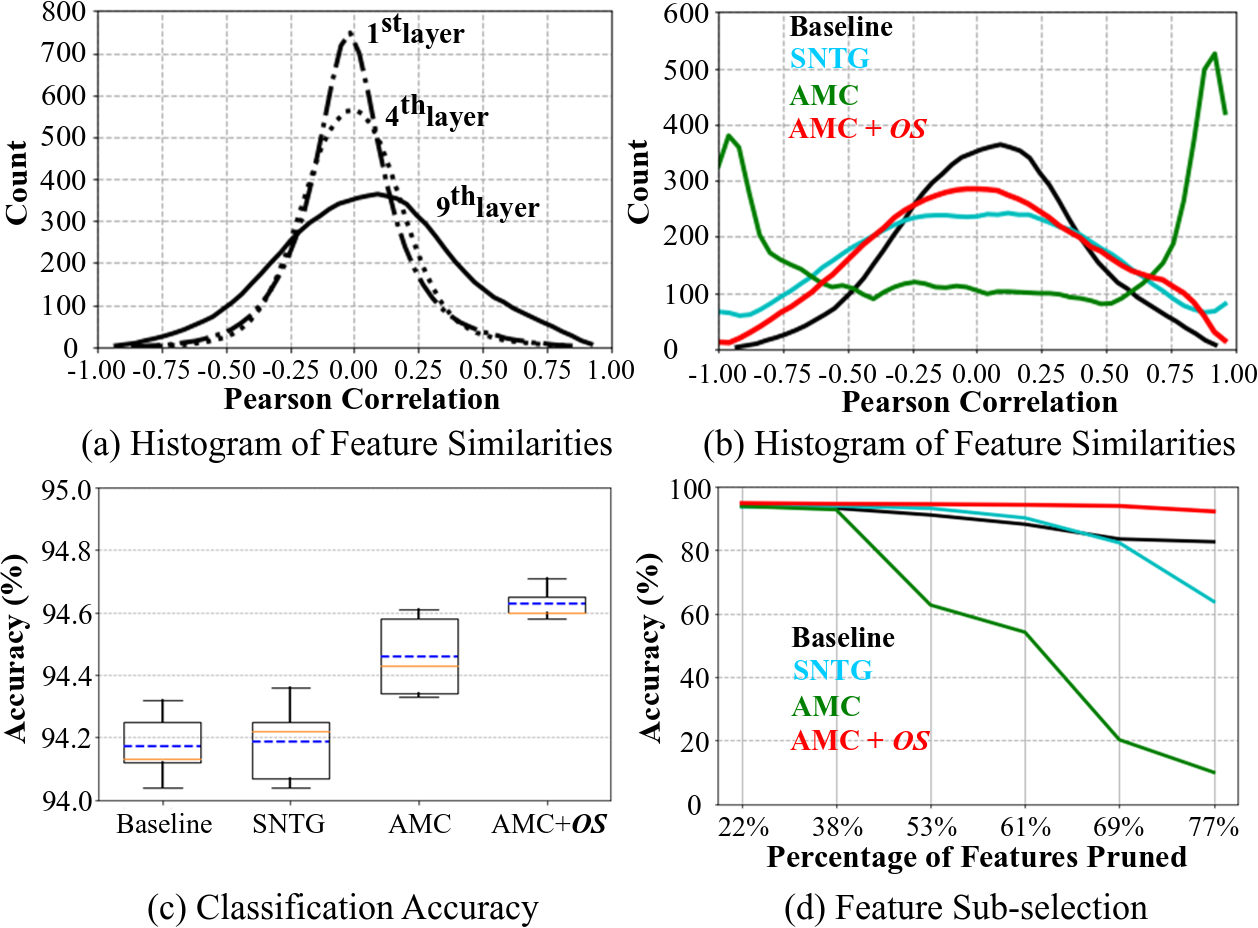}
	\end{center}
	\caption{Using the CIFAR10 dataset  \cite{krizhevsky2009learning} (a) Histogram of pairwise feature similarity of different layers in the Baseline $\Pi$ model \cite{laine2016temporal}. Higher correlations are observed for deeper layers in the model. (b) Feature similarity histograms of the $9^{\text{th}}$ layer for different approaches. We observe greater feature correlation for the contrastive techniques (SNTG, AMC). However, the proposed {\em OS} regularizer helps learn more decorrelated features. (c) Classification accuracy averaged over $5$ runs. (d) Sensitivity to pruning of feature maps in the $9^{\text{th}}$ layer.}     
	
	\label{fig:summary of result}
	\vspace{-0.2in}
\end{figure}

\begin{figure}[t!]
	\begin{center}
		\includegraphics[width=.75\linewidth]{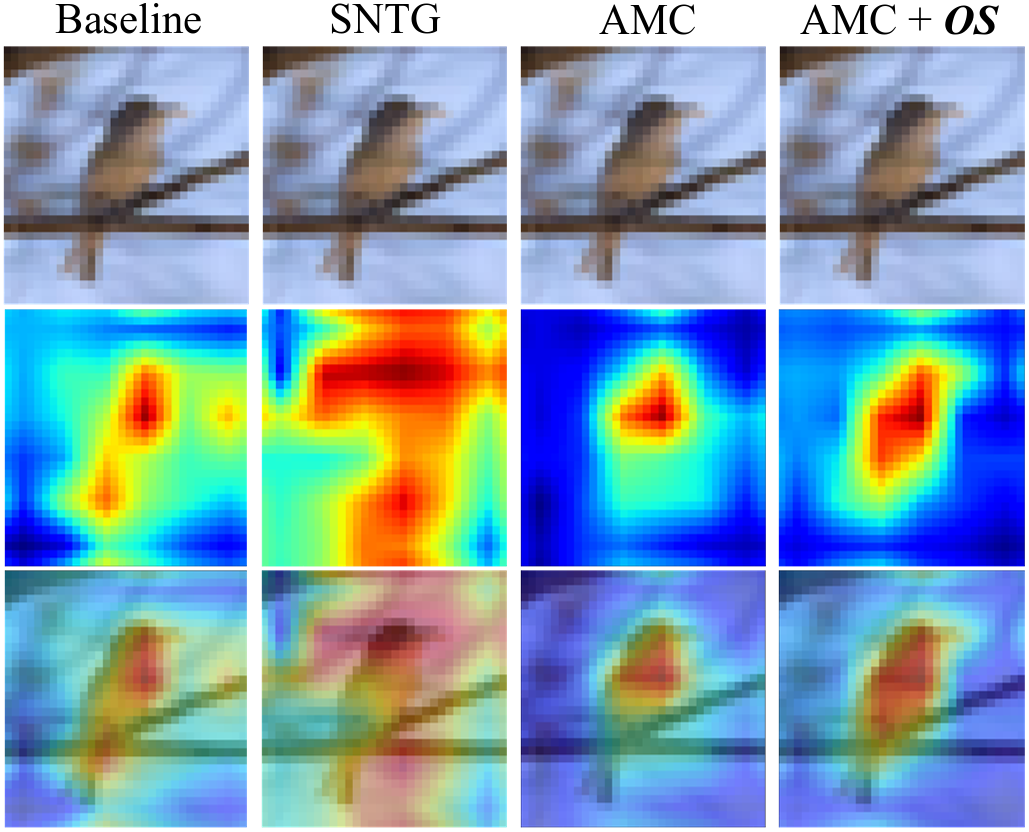}
	\end{center}
	\caption{Grad-CAM visualization (second row) of the baseline $\Pi$ model and different regularization approaches. The third row displays the overlaid image of the activation map on the input bird image. Applying the \emph{OS} constraint provides a more broad region of interest pertaining to the object of interest while focusing less on the background.}
	\label{fig:summary_activation_maps}
	\vspace{-0.1in}
\end{figure}

Deep learning models, specifically convolutional neural networks (CNNs) have become the default choice for image classification tasks. When training CNNs, the weights of the model are learnt using the categorical-cross-entropy loss. This loss function alone is enough for achieving high classification performance. However, the final model obtained exhibits certain unfavorable properties that hamper its overall reliability. For example, features learnt in the deeper layers of the network are highly correlated as seen in Figure \ref{fig:summary of result} (a). This makes the model sensitive to pruning techniques  (used for making smaller and compact models for edge applications) thereby causing a drop in the model's classification performance. The deep learning models also do not offer clear insight as to which parts of the image contribute towards the final label prediction. This is illustrated using the Grad-CAM \cite{selvaraju2017grad} visualization of the Baseline model in Figure \ref{fig:summary_activation_maps}. Also, since the ground-truth labels are represented as one-hot-coded vectors, the cross-entropy loss has the tendency to make the model overconfident since it only focuses on the predicted probability corresponding to the ground-truth label \cite{thulasidasan2019mixup}. Recent works try to address some of these challenges by using different regularization functions along with the cross-entropy loss \cite{luo2018smooth,choi2020amc}. However, these methods are not successful at addressing all of the challenges mentioned above.

Our approach is weakly motivated from image-formation physics and recent interest shown towards constraining latent variables of deep learning models with physics-based knowledge. The core idea behind our regularization function is simple. Many important physical factors like lighting, motion, pose have certain natural intrinsically non-Euclidean parameterizations in terms of geometric concepts like rotation groups, Grassmannian manifolds, and diffeomorphism groups. Under certain relaxed conditions, each of these factors can be embedded into a larger dimensional hypersphere, with differing dimensions. Furthermore, using orthogonality as a proxy for statistical independence, one arrives at an orthogonal-spherical model. One can further relax these constraints with fixed-size blocks per factor, resulting in simpler orthogonality constraints. This can be written down in closed form as a simple orthonormality term.

Our implementation of the regularization function is similar to \cite{shukla2019product} orthogonal-spheres technique which was first proposed in the context of disentangling auto-encoders. Imposing the orthogonal sphere constraint is much simpler than other complicated physical models \cite{balestriero2018mad, balestriero2018spline, bansal2018can, xie2017all, rodriguez2016regularizing}, thereby making it fairly general, flexible, and extensible. Our regularization function offers the following benefits: a) rich and diverse feature representations, b) robustness to pruning or feature sub-selection, c)  better semantic localization in the Grad-CAM class activation maps, and d) reduction in model calibration error.

As a preview of the benefits, Figure \ref{fig:summary of result} (b) shows the pairwise feature similarity histogram of the baseline, auxiliary regularization methods \cite{luo2018smooth,choi2020amc} and the proposed \emph{OS} regularizer. When compared to the AMC approach \cite{choi2020amc}, the \emph{OS} regularizer along with AMC helps learn more decorrelated features at the deeper layers of the network. Figure \ref{fig:summary of result} (c) shows marginal improvement in the overall classification performance on the CIFAR10 dataset \cite{krizhevsky2009learning}. In Figure \ref{fig:summary of result} (d) we observe that the proposed regularizer helps achieve greater robustness to pruning/feature sub-selection processes. Even with pruning rates as high as 77$\%$, we observe significantly higher classification performance when compared to the other methods. This helps minimize the learnt feature redundancy and reduces the need for model re-training after pruning. The {\em OS} regularizer can also result in a more plausible explanation in terms of semantically interpretable Grad-CAM activation maps as seen in Figure \ref{fig:summary_activation_maps}. While this interpretation does not always elucidate clearly how a given deep model works, they might nonetheless confer perceived trustworthiness in deep learning models for an end-user without affecting predictive performance \cite{ribeiro2016should}. Our method results in higher activation over the entirety of the target object while reducing the effect of the background. We feel this is because the {\em OS} regularization can produce more expressive feature representations by decorrelating across feature maps. As will be shown later, besides qualitative visualization, we also observe improved calibration metrics for better model interpretability and reliability on four benchmark datasets. 

\textbf{Paper Outline:} Section \ref{sec:Related_Work} discusses related work and background. Section \ref{sec:Proposed_Method} provides description of the proposed \emph{OS} regularization function. Section \ref{sec:Experiments} describes the experiments and results. Section \ref{sec:Conclusion} concludes the paper.

\section{Background and Related Work}\label{sec:Related_Work}
Here, we provide the necessary background and discuss related work that utilizes similar orthogonality type of regularization when training deep learning models. 


\subsection{Perturbation-based Method}\label{sec:perturbtion_method} 

For most of the experiments described in Section \ref{sec:Experiments} we use the $\Pi$ model \cite{laine2016temporal} as our baseline model. We evaluate the baseline model and its different variants under supervised and semi-supervised learning settings. For the semi-supervised setting, let us consider the training set to have $N$ samples, out of which we have label information for only $L$ samples. Let the labeled and unlabeled sets be denoted as $\mathcal{L}=\{(x_{i},y_{i})\}_{i=1}^{L}$ and $\mathcal{U}=\{x_{i}\}_{i=L+1}^{N}$ respectively. Here, $x_{i}\in X$ represents the input variables and $y_{i}\in {Y}=\{1,2,...,K\}$ represents the output variables with $K$ classes. The objective of the $\Pi$ model is to learn a function $f: {X}\rightarrow [0,1]^{K}$ with parameters $\theta\in \Theta$ by solving the below optimization problem

\begin{align}\label{eq:generic_equation}
L_{\Pi}=\min_{\theta}\sum_{i=1}^{L}L_\text{C}(f(x_{i};\theta),y_{i}) + \lambda R_\text{C}(\theta,\mathcal{L},\mathcal{U}) \\
R_\text{C}(\theta, \mathcal{L}, \mathcal{U}) = \sum_{i=1}^{L}\mathbb{E}_{\xi^{'},\xi}d(\Tilde{f}(x_{i};\theta^{'},\xi^{'}),f(x_{i};\theta,\xi))\label{eq:R_C}
\end{align}

Here, $L_\text{C}$ is the regular categorical-cross-entropy loss \cite{hinton2012improving, he2016deep, he2015delving} with $f(x;\theta)$ representing the predicted probability distribution $p(y|x;\theta)$. $R_\text{C}$ is a regularization term that is used to leverage the unlabeled training set. It is especially useful when working with very limited labeled training data, \emph{i.e.}, $L\ll N$. $R_\text{C}$ is added to the cross-entropy loss using a non-negative parameter $\lambda$. In $R_\text{C}$, $\Tilde{f}$ is a noisy teacher model with parameter $\theta^{'}$ that is subjected to random perturbations $\xi^{'}$. Similarly, $f$ is considered the student model that is parameterized by $\theta$ and is subjected to perturbations $\xi$. The teacher model is defined as an implicit ensemble of previous student models. The student model is expected to predict consistently with $f(x;\theta)$. $d$ captures the divergence between the two predicted probability distributions. The $\Pi$ model feeds each unlabeled sample into the classifier twice. Each time it does so with a different dropout, noise, and augmentation such as random translation. 

\subsection{SNTG and AMC Regularization}\label{sec:sntg_amc} 
The main drawback of the $\Pi$ model is that it only penalizes inconsistent predictions of the unlabeled data under different perturbations while ignoring the relationships between the neighboring samples. To address this issue \cite{luo2018smooth} proposed the Smooth Neighbors on Teacher Graphs (SNTG) regularizer as shown below
\begin{equation}\label{eq:contrastive_loss}
L_\text{SNTG} = 
\begin{cases}
\left\|l_{i}-l_{j}\right\|^{2} & \text{if } S_{ij}=1 \\
\max(0, m_{e}-\left\|l_{i}-l_{j}\right\|)^{2} & \text{if } S_{ij}=0.
\end{cases}
\end{equation}

Here, $l_i, l_j$ represent low-dimensional feature representations (obtained as output from a specific layer in the model) of the corresponding input images; $m_e>0$ is a pre-defined Euclidean margin; $\left\| \cdot \right\|$ denotes the Euclidean distance and $S_{ij}$ represents the similarity matrix. The similarity matrix is assigned a value 0 if the predicted labels from the teacher model for a given pair of samples are different (non-neighboring pairs), otherwise it is assigned a value 1 (neighboring pairs). Using this loss function the neighboring samples are encouraged to minimize their distances while the non-neighboring samples are pushed apart from each other while maintaining a minimum distance of $m_{e}$.  

Another recent work by \cite{choi2020amc} proposed the Angular Margin Contrastive (AMC) regularizer that simply replaces the Euclidean metric defined in Equation \ref{eq:contrastive_loss} with a geodesic metric defined for unit-normalized latent representations. This helps improve the classification performance and contributes towards better explainable interpretations in terms of Grad-CAM class activation maps. They proposed the following regularizer

\begin{align}\label{eq:AMC-Loss}
L_\text{AMC} = 
\begin{cases}
(\cos^{-1}\langle\,z_{i},z_{j}\rangle)^{2} & \text{if } S_{ij}=1 \\
\max(0, m_{g}-\cos^{-1}\langle\,z_{i},z_{j}\rangle)^{2} & \text{if } S_{ij}=0.
\end{cases}
\end{align}

Here, $m_g>0$ represents the angular margin and $z_i$ is just a unit-normalized representation of $l_i$ in Equation \ref{eq:contrastive_loss}.

\subsection{Constrained Neural Networks}\label{sec:orthogonality_const} 

Several recent works propose constraining or regularizing different parts of the deep learning model to incorporate different desirable properties. For example, constraining deep networks has shown to help achieve better training stability and performance gain \cite{balestriero2018mad, balestriero2018spline, bansal2018can, xie2017all}; preserves the energy of the model and make uniform spectra \cite{zhou2006special}; stabilize the distribution of activation maps in CNNs \cite{rodriguez2016regularizing} and avoids exploding/vanishing gradient issues \cite{arjovsky2016unitary}. The following two approaches incorporate the concept of orthogonality in deep neural networks: 1. Kernel Orthogonality \cite{balestriero2018mad, balestriero2018spline, bansal2018can, xie2017all};
2. Orthogonal CNNs \cite{wang2020orthogonal}. These methods constrain the learnable weights in specific layers, to satisfy certain orthogonality conditions. Whereas, in our method, we constrain the outputs of the layers, i.e. the features themselves. This is motivated by the reasoning that deep CNNs learn representations that can be seen as a complex interaction between various physical factors but not necessarily related to basic constraints from image-formation. In this regard, this process could be improved further by directly imposing constraints on deep features, making them decorrelated.    

\section{Orthogonal Sphere (\emph{OS}) Regularization}\label{sec:Proposed_Method}

As mentioned earlier, CNN models trained using just the regular cross-entropy loss automatically encode different complex interactions between various physical factors like lighting and pose. However, the features learnt are not necessarily constrained or related to basic constraints from image-formation. They also end up being highly correlated especially in the deeper layers of the network. This results in feature redundancy with the final model being highly sensitive to pruning techniques. Regularization functions when used along with the cross-entropy loss can help address some of these challenges. In this section, we describe the proposed \emph{OS} regularization that helps incorporate different desirable properties automatically without generating any additional learnable parameters. 

\begin{figure}[t!]
	\begin{center}
		\includegraphics[width=0.99\linewidth]{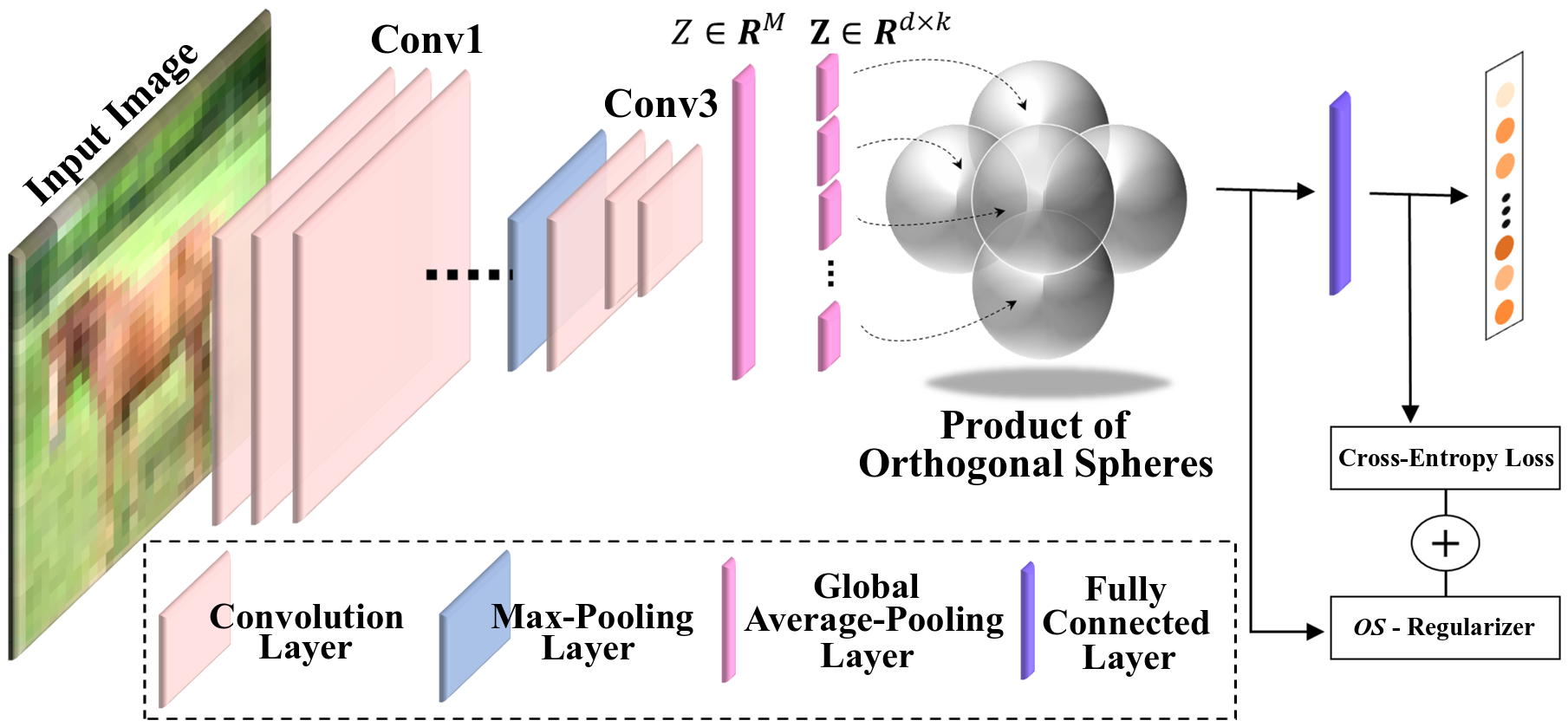}
	\end{center}
	\caption{Illustration of the implementation of the proposed \emph{OS} regularizer. Here, the \emph{OS} regularization is applied to the global-average-pooling layer's output. However, the proposed constraint can be easily applied to the output of any layer in the deep neural network.}
	\label{fig:pipeline}
\end{figure}

The framework of the proposed regularizer is illustrated in Figure \ref{fig:pipeline}. For a given input image, $Z\in \mathbb{R}^{M}$ represents the global-average-pooled output of a specific layer in the CNN (\textbf{Conv3} in the illustrated example). Latent representation $Z$ can be further converted into separate feature blocks given by $\mathbf{Z}\in\mathbb{R}^{d\times k} = [\mathbf{z}^1,\mathbf{z}^2,\dots,\mathbf{z}^k]$. Here, $k$ represents the number of partitions and $d$ is the length/dimension of each partition. Depending on what we set these values to, $\mathbf{Z}$ can either be a fat matrix ($d<k$) or a tall matrix ($d>k$). In either case, we want to regularize the off-diagonal elements in the matrix to be zero to make it as orthogonal as possible. The condition for orthogonality is defined below
\begin{equation}\label{eq:orthogonal_condition}
\langle \mathbf{z}^{i},\mathbf{z}^{i'}\rangle = 
\begin{cases}
1, & \text{if } i=i',\\
0, & \text{otherwise.}
\end{cases}
\end{equation}

We can apply this condition on matrix $\mathbf{Z}$ in closed form as a simple orthonormality term as shown below

\begin{align}\label{eq:OS-Loss}
L_\text{OS} &= \left\|\mathbf{Z}^{\top}\mathbf{Z}-\mathbf{I}\right\|^{2}_{F}.
\end{align}

Here, $L_\text{OS}$ is our \emph{OS} regularizer, with $\left\|\cdot\right\|_{F}$ representing the Frobenius norm and $\mathbf{I}$ being the $k\times k$ identity matrix. This function can be used together with the cross-entropy loss defined for the $\Pi$ model (described in Equation \ref{eq:generic_equation}) and/or with the other auxiliary loss functions (defined in Equations \ref{eq:contrastive_loss} and \ref{eq:AMC-Loss}) as shown below

\begin{align}\label{eq:total_loss}
L_\text{Total} &= L_{\Pi} + \lambda_{1}L_\text{Aux} + \lambda_{2}L_\text{OS}.
\end{align}

\begin{algorithm}[t!]
	\caption{{Mini-batch training of the \emph{OS} regularizer.} \label{Alg:mini-batch_training}}
	\textbf{Input: } $(x_i,y_i)$: input and output variables in labelled set $\mathcal{L}$. \\
	\textbf{Require: } $\mathbf{Z}_{i}$ = stacked matrix of the latent code vectors. \\
	\textbf{Require: } $w(t)$ = unsupervised weight ramp-up function. \\
	\textbf{Require: } $f_{\theta}(x)$ = CNN with parameter $\theta$. \\
	\begin{algorithmic}[1]
		\For{$t$ in $[1, \text{num\_epochs}]$}
		\For{each minibatch $B$}
		\State $f_{i}$ $\leftarrow$ $f_{\theta}(x_{i}\in B)$ evaluate network outputs
		\State $\tilde{f}_{i}$ $\leftarrow$ $f(x_{i}\in B)$ given by the teacher model
		\For{$(x_{i}, x_{j})$ in a mini-batch pairs $S$ from $B$}
		\State compute $S_{ij}$ based on target prediction. 
		\EndFor
		\State loss $\leftarrow$ $-\frac{1}{|B|}\sum_{i\in (B\cap \mathcal{L})}\log[f_{i}]_{y_{i}}$ 
		\State \qquad\quad $+ w(t)[\lambda\frac{1}{|B|}\sum_{i\in B}d(\tilde{f}_{i},f_{i})$
		\State \qquad\quad $+ \lambda_{1}\frac{1}{|S|}\sum_{i,j\in S}(L_\text{SNTG} \text{ or } L_\text{AMC})$
		\State \qquad\quad $+ \lambda_{2}\frac{1}{|B|}\sum_{i\in B}L_\text{OS}(\mathbf{Z}_{i})]$
		\State update $\theta$ using Adam optimizer.
		\EndFor
		\EndFor
		\State return $\theta$
	\end{algorithmic}
\end{algorithm}

The pseudo-code is presented in Algorithm \ref{Alg:mini-batch_training}. For our experiments in Section \ref{sec:Experiments} we also explore normalizing the latent representation $Z$ by projecting it onto a unit-hypersphere of radius $s$. Through our empirical experiments, we find that this normalization step helps learn more semantically meaningful Grad-CAM class activation maps. Whereas, the original unnormalized latent representation facilitates better utilization of the smaller labeled set $\mathcal{L}$. In Figure \ref{fig:pipeline}, the proposed \emph{OS} regularizer is shown to be applied to the output representations of the global-average-pool layer. However, it can easily be applied to the flattened output of any other layer in the deep neural network by taking the average of all $h.w$ values to reduce $h\times w$ feature map to a single number where $h$ and $w$ represent the height and width of the feature map in each layer. We demonstrate this through the different experiments shown in Section \ref{sec:Experiments}.

\begin{figure}[t!]
	\centering
	\vspace{-0.05in}
	\includegraphics[width=0.99\linewidth]{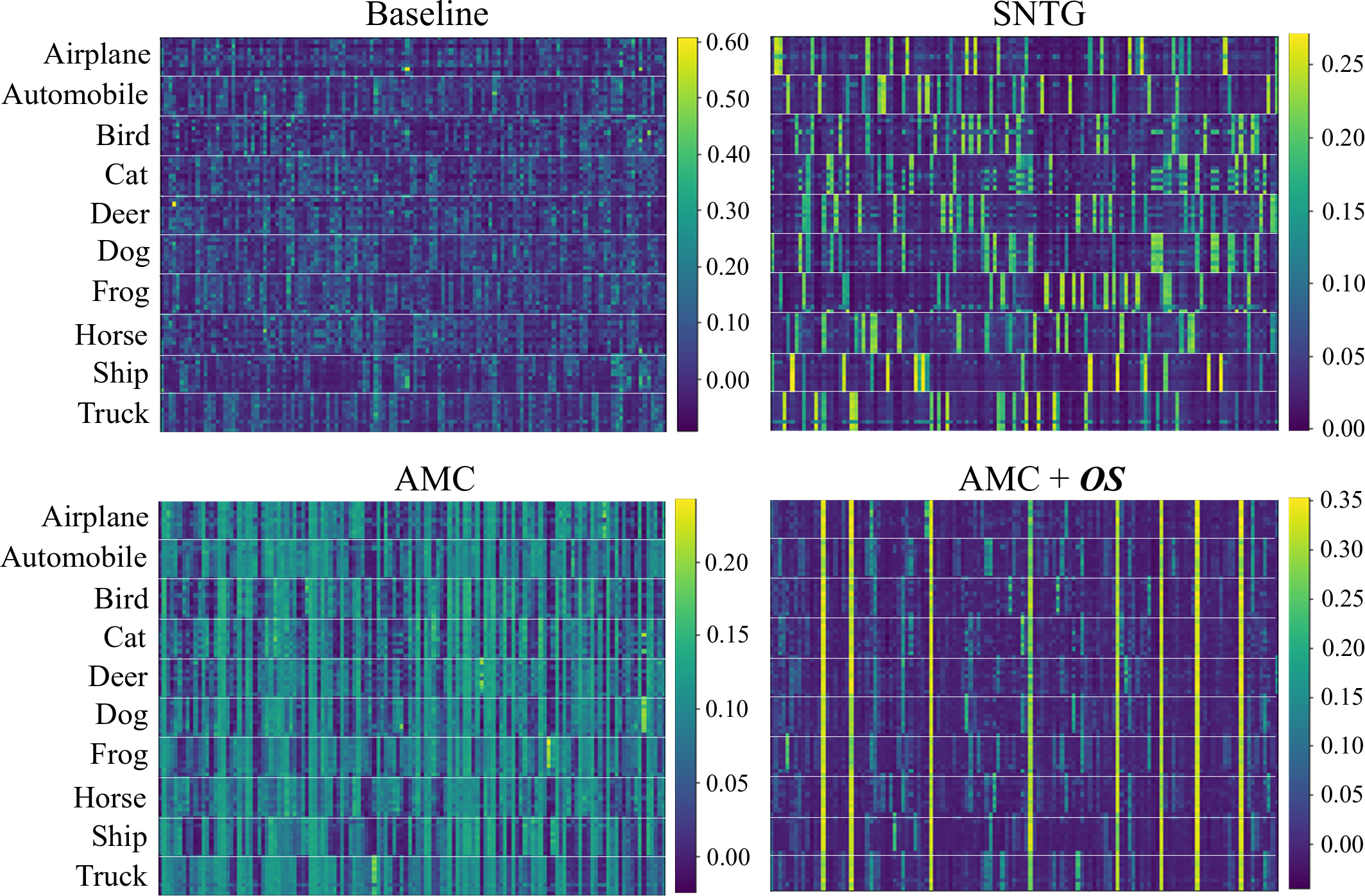}
	\caption{Visualization of the latent representation $Z$ for different loss functions. Latent representations were extracted for 10 random test images per class from the CIFAR10 dataset and stack them to get a matrix of size $100\times128$. Best viewed in color.}
	\label{fig:penultimate_layer}
\end{figure}

Figure \ref{fig:penultimate_layer} illustrates how the latent representation $Z$ looks like for different test images in CIFAR10 when using different loss functions. Using just the baseline $\Pi$ model, the feature values are spread over a wider range and it is also difficult to observe patterns unique to each label category. Using regularization methods like SNTG and AMC helps reduce the range of feature values. However, even with these regularizers, we are still unable to learn unique and sparse feature representations. An interesting byproduct of the \emph{OS} regularizer (when used with AMC) is that the features learnt are more sparse and distinct across the different classes while still sharing sufficient similarities within the same class.

\begin{table*}[t!]
	\centering
	\caption{Classification results averaged over 5 runs. All results were reported using $k=16$ (number of latent feature blocks).}\label{benchmark_result}
	\scalebox{0.72}{\begin{tabular}{ |c|c|c|c|c|c|c|c|c|c| } 
			\hline
			\multirow{2}{*}{\textbf{Method}} & \multicolumn{2}{c|}{\textbf{CIFAR10}} & \multicolumn{3}{c|}{\textbf{SVHN}} & \multicolumn{2}{c|}{\textbf{CIFAR100}} & \multicolumn{1}{c|}{\textbf{tiny ImageNet}} \\ [0.5ex] 
			\cline{2-9}
			& 4000 & All labels & 500 &  1000 & All labels & 10000 & All labels & All labels \\ [0.5ex] 
			\hline
			$\Pi$ model & 86.12$\pm$0.28 & 94.17$\pm$0.09 & 91.60$\pm$0.05 & 94.57$\pm$0.07 & 97.37$\pm$0.02 & 58.42$\pm$0.32 & 72.82$\pm$0.16 & 55.31\\ [0.5ex] 
			$\Pi$ model + {\em \textbf{OS}} & \textbf{86.86$\pm$0.11}* & \textbf{94.29$\pm$0.14} & - & - & - & - & - & -\\ [0.5ex]
			$\Pi$ model + SNTG  & 87.45$\pm$0.13 & 94.18$\pm$0.11 & 95.14$\pm$0.04 & 95.87$\pm$0.05 & 97.49$\pm$0.02 & 59.45$\pm$0.36 & 72.99$\pm$0.10 & -\\ [0.5ex] 
			$\Pi$ model + SNTG + {\em \textbf{OS}} &  \textbf{87.84$\pm$0.19}* & \textbf{94.28$\pm$0.08} & - & - & - & - & - & - \\ [0.5ex] 
			$\Pi$ model + AMC & 87.11$\pm$0.12 & 94.45$\pm$0.11 & 93.31$\pm$0.07 & 94.93$\pm$0.07 & 97.46$\pm$0.03 & 58.73$\pm$0.39 & 73.13$\pm$0.18 & 55.58\\ [0.5ex] 
			$\Pi$ model + AMC + {\em \textbf{OS}} & \textbf{87.44$\pm$0.09}* & \textbf{94.62$\pm$0.04} & \textbf{93.77$\pm$0.08} & \textbf{95.35$\pm$0.08} & \textbf{97.51$\pm$0.04} & \textbf{59.71$\pm$0.16}* & \textbf{74.03$\pm$0.12}* & \textbf{56.65}\\ [0.5ex] 
			\hline
	\end{tabular}}
\end{table*}

\section{Experiments}\label{sec:Experiments}

\noindent\textbf{Dataset Description:} We conduct experiments using the following four image datasets. 

\begin{itemize}
	\item \textbf{CIFAR10 \cite{krizhevsky2009learning}:} Contains color images of $10$ label categories, where each image has a size of size $32\times32$. Each category has 5000 training images and 1000 test images.
	\item \textbf{CIFAR100 \cite{krizhevsky2009learning}:} Contains color images of $10$ label categories, where each image has a size of size $32\times32$. Each category has 500 training images and 100 test images.
	\item \textbf{SVHN \cite{netzer2011reading}:} Contains color images of $10$ label categories, where each image has a size of size $32\times32$. Each category has 73257 training images and 26032 test images.
	\item \textbf{Tiny-ImageNet \cite{deng2009imagenet}:} Contains color images of $200$ label categories, where each image has a size of size $64\times64$. Each category has 500 training images, 50 validation images and 50 test images.
\end{itemize}

\noindent\textbf{Training Protocol and Hyper-parameter Settings:} All our experiments were conducted using a single GeForce RTX 2080 Ti GPU and Python $3.7$ with Tensorflow $1.12.0$. For most datasets, our models were trained for $300$ epochs using the Adam optimizer,  batch-size = $100$ and minimum-learning-rate = $0.003$ except for tiny-ImageNet trained the models for $200$ epochs. We used the default Adam momentum parameters: $\beta_{1}=0.9$ and $\beta_{2}=0.999$. Following \cite{laine2016temporal}, we ramp up the weight parameter $w(t)$ and learning rate for the first $80$ epochs with weight $w(t)=\exp[-5(1-\frac{t}{80})^{2}]$ and ramp down during the last 50 epochs. The ramp-down function is defined as $\exp[-12.5(1-\frac{300-t}{50})^{2}]$. As studied in \cite{laine2016temporal, luo2018smooth, choi2020amc}, we set $\lambda_{1}=0.4$, $m_{e}=1.0$ for SNTG and use $\lambda_{1}=0.1$, $m_{g}=0.5$ for AMC with the balancing parameter $\lambda$ being $1$ for all models. We apply batch-normalization with leaky ReLu activation \cite{maas2013rectifier} with $\alpha=0.1$. Please note, for the proposed \emph{OS} regularization, reported results without mark(*) represent normalized features that were rescaled by $s=3$. Also, the second balancing coefficient $\lambda_{2}$ is set to $7e^{-5}$ for unmarked results and $5e^{-4}$ for the marked(*) ones. 

\subsection{Image Classification on Benchmark Datasets}\label{sec:classification_results} 

For CIFAR10, we compare the image classification performance of the proposed \emph{OS} regularization when used with the baseline $\Pi$ model and with each of the other regularization functions. The classification results are evaluated under semi-supervised and supervised learning settings and summarized in Table \ref{benchmark_result}. The \emph{OS} regularizer helps marginally improve the overall classification performance for each case explored. For SVHN, CIFAR100 and tiny-ImageNet we only combine the \emph{OS} regularization with the AMC approach. However, the $\Pi \text{ model} +AMC+\emph{OS}$ method still performs better than all other cases for each dataset. Specifically, we see a greater improvement for datasets that have more number of classes (CIFAR100, tine-ImageNet) using the proposed approach. Please note, we apply the \emph{OS} regularization on each output of the last set of the convolutional layer (\textbf{Conv3} illustrated in Figure \ref{sec:Proposed_Method}) and the global-average pooled output in the classification model. We do this to address the issue of deeper layers in the network learning more correlated features. We want to minimize the number of correlated features learnt by applying the proposed regularization function. For all these cases we set $k=16$ for the \emph{OS} regularizer. Table \ref{number of partitions} shows the robustness in classification performance of the \emph{OS} regularizer for different values of $k$. The robustness is evaluated using the CIFAR10 and SVHN datasets.


\noindent\textbf{Comparison with other orthogonality-based methods:} Next we compare the classification performance of our {\em OS} regularization against the Kernel-Orth \cite{xie2017all} and OCNN \cite{wang2020orthogonal} methods. For this experiment we used the ResNet18 and ResNet34 as our baseline classification models. We use these models to be able to compare against the reported performance of the Kerenl-Orth and OCNN methods. We apply the proposed \emph{OS} regularizer at the output of each residual block and at the last fully connected layer. We set $\lambda = 1e^{-4}$ and $k=16$. Table \ref{comparison_ocnn} displays the top-1 classification accuracy on CIFAR100 along with the total time taken for training each model. The \emph{OS} regularizer performs better than the individual baseline ResNet models and the Kernel-Orth method. Our method's performance is slightly lower than OCNN. However, \emph{OS} helps speed up the overall time taken to train the model.

\begin{table}[t!]
	\centering
	\caption{Image classification results of the \emph{OS} regularization on CIFAR10 and SVHN with different $k$ (number of latent block partitions). The results are averaged over 5 runs.}\label{number of partitions}
	\scalebox{0.725}{\begin{tabular}{ |c|c|c|c|c| } 
			\hline 
			& $\Pi$ model & + \emph{OS} (\textbf{$k=16$}) & + \emph{OS} (\textbf{$k=32$}) & + \emph{OS} (\textbf{$k=64$})   \\ [0.5ex] 
			\hline
			CIFAR10 & 94.17$\pm$0.09 & \textbf{94.62$\pm$0.04} & \textbf{94.52$\pm$0.08} &  \textbf{94.59$\pm$0.11}\\ [1ex]
			SVHN & 97.37$\pm$0.02 & \textbf{97.54$\pm$0.02}* &  \textbf{97.53$\pm$0.03}* & \textbf{97.52$\pm$0.02}*\\ [1ex]
			\hline
	\end{tabular}}
	
\end{table}

\begin{table}[t!]
	\centering
	\caption{Comparison of different orthogonality-based approaches on CIFAR100 using ResNet18 or ResNet34  as the baseline model.}\label{comparison_ocnn}
	\scalebox{0.85}{\begin{tabular}{ |c|c|c|c|c| } 
			\hline
			Method & ResNet18 & ResNet34 & Total training time\\ 
			\hline
			Baseline & 75.3 & 76.7 & 2$h$ 11$m$\\ 
			+ Kernel-Orth & 76.5 & 77.5 & - \\ 
			+ OCNN & 78.1 & 78.7 & 3$h$ 22$m$\\ 
			+ {\em \textbf{OS}} & \textbf{77.22} & \textbf{78.42} & 2$h$ 47$m$\\ 
			\hline
	\end{tabular}}
\end{table}


\begin{figure*}[h!]
	\begin{center}
		\includegraphics[width=0.80\linewidth]{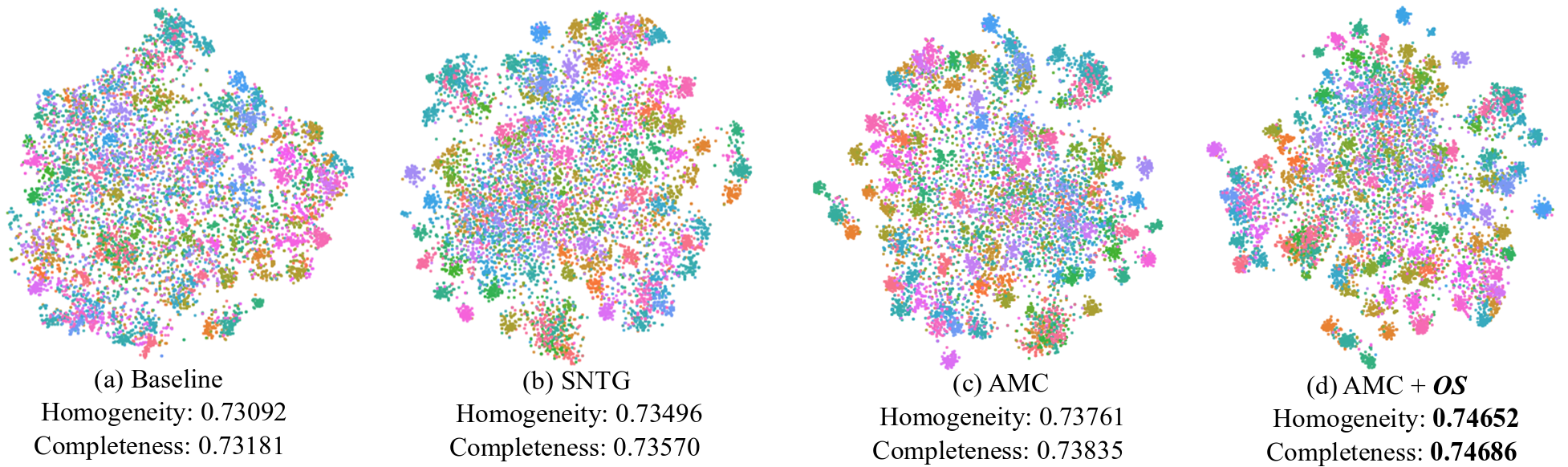}
	\end{center}
	\caption{Two-dimensional t-SNE plots of the penultimate layer's output for (a) Baseline $\Pi$ model, (b) SNTG and (c) AMC and (d) AMC+\emph{OS}. Each color denotes a different class. The representation learnt using the {\em OS} regularization promotes better inter-class separability and intra-class compactness.}
	\label{fig:tSNE}
	\vspace{-0.2in}
\end{figure*}


\subsection{t-SNE Visualization and Cluster Metrics}\label{sec:feature_maps}

Here we plot 2D t-SNE \cite{maaten2008visualizing} visualizations of the penultimate layer's output representations on CIFAR100 as seen in Figure \ref{fig:tSNE}. We observe that using a regularization function definitely helps get better t-SNE visualizations. In addition, our \emph{OS} method helps get tighter clusters and better boundary separation between the different classes. Our observations of the t-SNE plots are further backed by the Homogeneity and Completeness scores shown in Figure \ref{fig:tSNE}. These metrics help quantify the quality of the clusters, with a higher score indicating a better cluster representation.



\subsection{Feature Similarity, Calibration and Grad-CAM Visualization}\label{sec:feature_similarity}

\begin{figure}[t!]
	\begin{center}
		\includegraphics[width=0.99\linewidth]{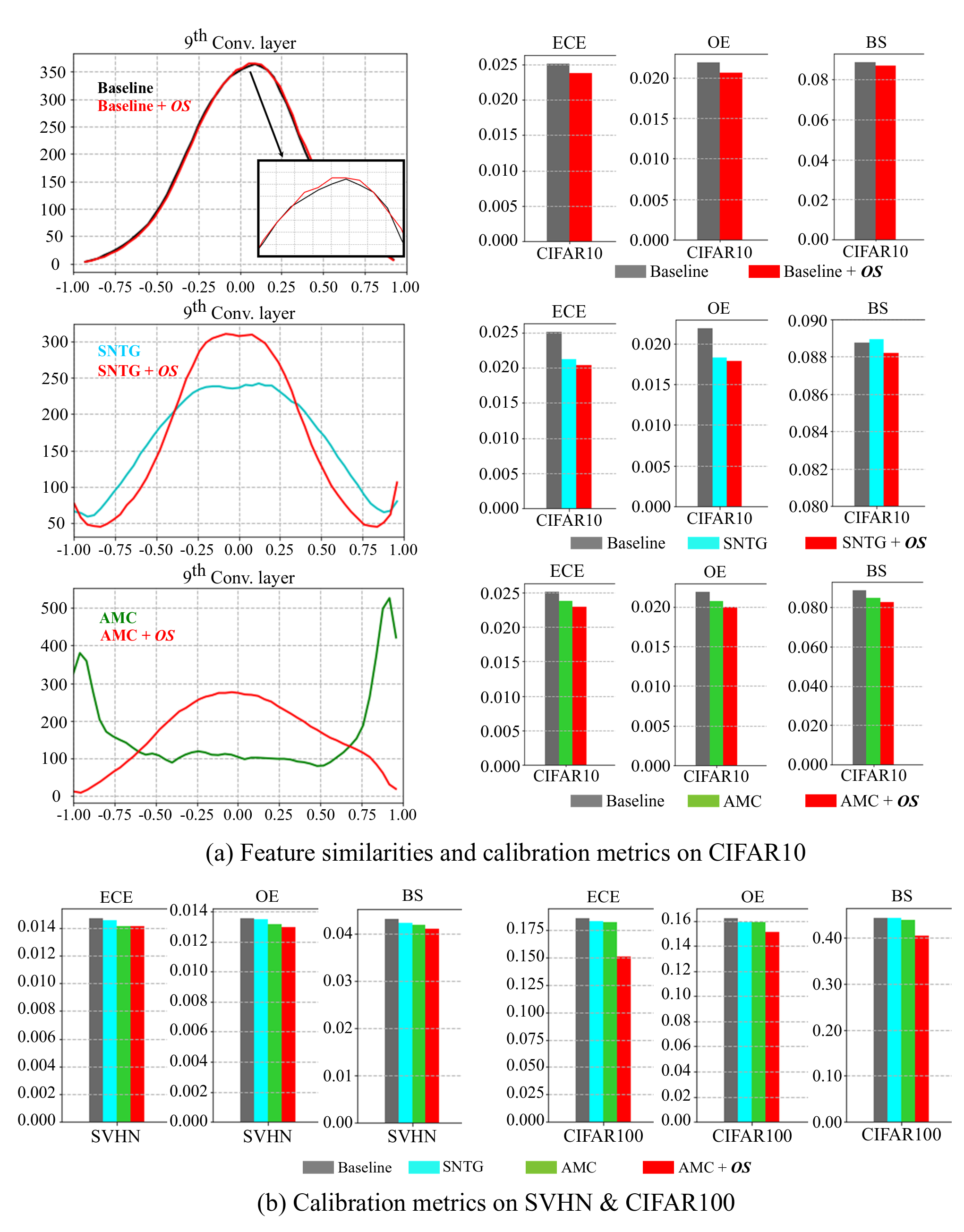}
	\end{center}
	\caption{Histogram of feature similarity along with the ECE, OE, and BS calibration scores on (a) CIFAR10, (b) SVHN and CIFAR100. The calibration metric scores reported were averaged for 5 models in each case.}
	\label{fig:feature_similar_calibration}
	\vspace{-0.2in}
\end{figure}

To investigate feature similarity, we use gradient information patterns of a given target label flowing into the convolutional layer with respect to the input image and combine feature maps using the gradient signal. This allows us to see which feature locations are being used the most for predicting the target label. 
Let us consider the output feature representations of a specific layer in the deep network model and denote it as $G\in\mathbb{R}^{m\times h\times w}$, where $m$ represents the number of output channels, and $h$ and $w$ represent the height and width of the output feature map. Furthermore, each output channel is flattened such that we can now represent $G\in\mathbb{R}^{m\times p}$, where $p$ is equal to the product of $h$ and $w$. Next, we compute a correlation matrix denoted as $\text{corr}(G)$. Each element in the matrix corresponds to the pearson correlation coefficient calculated for a given pair of channels from a set of $m$ channels. This results in a $m\times m$ matrix. To generate the feature-similarity histograms shown in Figures \ref{fig:summary of result}, \ref{fig:feature_similar_calibration} we only use the off-diagonal elements in the correlation matrix. As seen in Figure \ref{fig:feature_similar_calibration}, the \emph{OS} regularizer helps learn more decorrelated features when used with the baseline $\Pi$ model and the other regularization functions.


We also look at how well the models were calibrated using the proposed \emph{OS} regularizer. Calibration metrics help assess if the predicted softmax scores obtained from a deep neural network are food indicators of the actual likelihood of the correct predictions. We evaluate our models using the Expected Calibration Error (ECE), Overconfidence Error (OE), and Brier Score (BS) \cite{guo2017calibration, gneiting2007strictly}. Each of these calibration metrics is defined as follows
\begin{itemize}
	\item ECE = $\sum_{m=1}^{M}\frac{|B_{m}|}{N}|\text{acc}(B_{m})-\text{conf}(B_{m})|$
	\item OE = $\sum_{m=1}^{M}\frac{|B_{m}|}{N}[\text{conf}(B_{m})\times \text{max}(\text{conf}(B_{m})-\text{acc}(B_{m}),0)]$
	\item BS = $\frac{1}{N}\sum_{n=1}^{N}\sum_{k=1}^{K}[p_{\theta}(\hat{y}_{n}=k|x_{n})-\mathbbm{1}(y_{n}=k)]^{2}$
\end{itemize}

\begin{figure*}[t!]
	\begin{center}
		\includegraphics[width=0.95\linewidth]{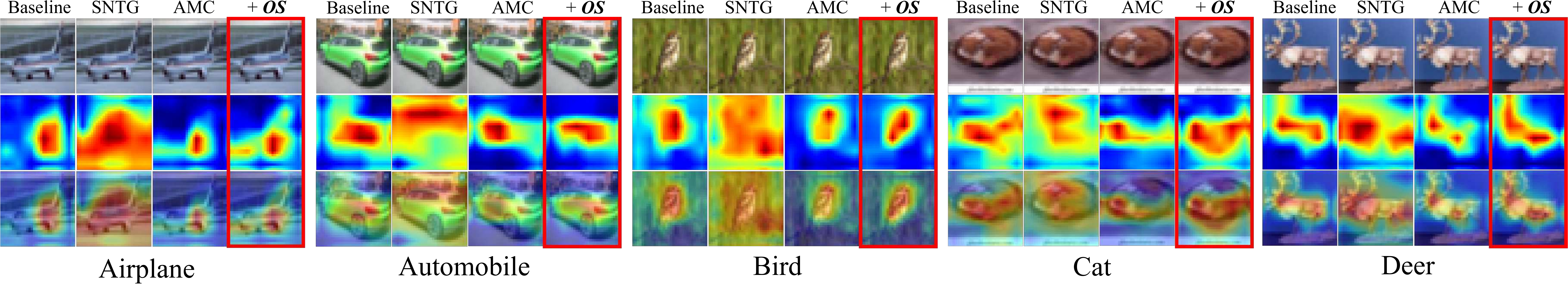}
	\end{center}
	\caption{Grad-CAM visualization of different test images in CIFAR10. The first row indicates the test image, second row corresponds to the computed Grad-CAM activation map for each method and third row represents the overlaid image.}
	
	\label{fig:Activation_Map_all}
	\vspace{-0.1in}
\end{figure*}

Here, $B_{m}$ represents the number of predictions falling in bin $m$ and $K$ is the number of classes. The accuracy of the model is denoted as $\text{acc}(B_{m})$ and the model's average confidence is denoted as $\text{conf}(B_{m})$. Figure \ref{fig:feature_similar_calibration} illustrates the calibration metric scores for the difference cases. Note, models with lower calibration scores are better. We observe lower calibration scores when we use the \emph{OS} regularizer with the baseline $\Pi$ model and the other regularizers. In Figure \ref{fig:Activation_Map_all} we also provide Grad-CAM visualizations of 5 test images in CIFAR10. The AMC+{\em OS} regularizer shows qualitatively better semantic interpretation, \emph{i.e.}, the objects of interest are widely captured while preserving distinct regions. However, the other approaches appear more spread out and less `on target'. By emphasizing important regions, the {\em OS} can bring out stronger explainable performance.


\subsection{Feature Sub-Selection / Pruning}\label{sec:Feature_selection}

In this section we study the effect of feature sub-selection or pruning the model. We only focus on the effect of pruning at the last convolutional layer ($9^{\text{th}}$ layer) of a model. We know that the features learnt by the deeper layers of a standalone model are generally more correlated. This makes the deeper parts of the network more sensitive to different pruning strategies. We wanted to study the effect of pruning on the overall classification performance of the model when we employ different regularization methods.

\begin{table}[t!]
	\centering
	\scalebox{0.725}{\begin{tabular}{|c|c|c|c|c|} \hline
			Prune-rate  & Baseline & SNTG & AMC & AMC + \textbf{\textit{OS}} \\[0.5ex] \hline 
			$\approx22\%$ & 93.82$\pm$0.14 & 93.60$\pm$0.15 & 93.99$\pm$0.17 & \textbf{94.65$\pm$0.21} \\ [0.5ex]
			$\approx38\%$ & 93.09$\pm$0.10 & 93.84$\pm$0.26 & 88.39$\pm$2.95 & \textbf{94.60$\pm$0.07} \\[0.5ex]
			$\approx53\%$ & 91.23$\pm$0.61 & 93.13$\pm$0.50 & 55.33$\pm$10.74 & \textbf{94.48$\pm$0.07} \\[0.5ex]
			$\approx61\%$ & 89.44$\pm$0.73 & 90.76$\pm$2.29 & 34.12$\pm$16.53 & \textbf{94.25$\pm$0.29} \\[0.5ex]
			$\approx69\%$ & 86.60$\pm$1.81 & 85.20$\pm$7.36 & 19.34$\pm$5.27 & \textbf{93.78$\pm$0.34} \\[0.5ex]
			$\approx77\%$ & 84.12$\pm$1.04 & 75.58$\pm$10.30 & 15.92$\pm$4.84 & \textbf{91.00$\pm$2.82} \\[0.5ex] \hline
	\end{tabular}}
	\vspace{0.05in}
	\caption{Average classification performance over $5$ trained models on the CIFAR10 dataset at different pruning rates. Note, we only prune the feature maps of the last convolution layer ($9^{\text{th}}$ layer) in the $\Pi$-model.}\label{table:ablation_table}
	\vspace{-0.100in}
\end{table}

We conduct this experiment using the CIFAR10 dataset. To determine which feature maps to prune in the $9^{\text{th}}$ layer, we measure the relative importance of the feature maps based on simple magnitude measurement. We do this in the following sequence of steps. First, we split the $10000$ test images into two equal sets, with one set serving as the validation set and the other as our test set. Second, we extract the $9^{\text{th}}$ layer feature maps for the validation set. For each image we get feature maps of dimension $6\times 6\times 128$, where $128$ denotes the number of output feature maps and $6$ denotes each feature map's height and width. Next, we compute the average of each feature map, thereby giving us a $128$-dimensional feature for each image. Thus, for the validation set we get a matrix of size $5000\times 128$. Finally, we compute the average across the $5000$ features thereby resulting in an array of size $128$. In this array we identify $n$ elements with the lowest magnitudes and discard the corresponding feature maps in the last convolution layer ($9^{\text{th}}$ layer). We define the pruning rate as $\frac{n\times100}{128}$, where $n$ denotes the number of feature maps pruned and $128$ is the total number of feature maps.

\begin{figure}[t!]
	\centering
	\includegraphics[width=0.75\linewidth]{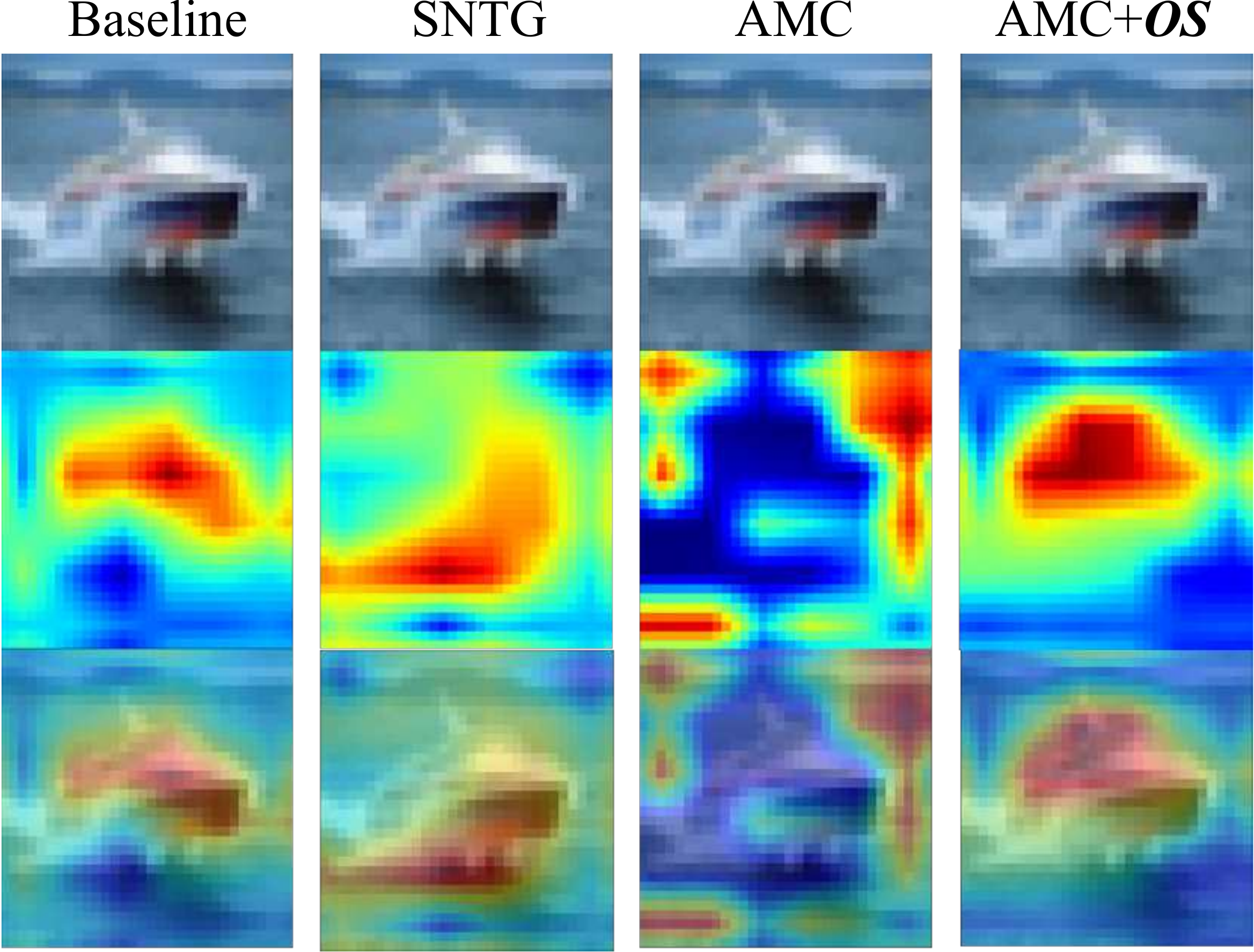}
	\vspace{0.05in}
	\caption{Grad-CAM visualization of models trained using different approaches when the last convolution layer is subjected to a pruning rate of $77\%$.}
	
	\label{fig:ablation_study}
	\vspace{-0.1in}
\end{figure}

Once we pruned the model we evaluate its performance using the 5000 test-set images. Table \ref{table:ablation_table} shows the classification performance of the different methods on the CIFAR10 dataset at different pruning rates. As the pruning rate increases, we notice that the classification performance drastically drops for the baseline $\Pi$ model, SNTG and AMC methods. Models trained using the proposed \emph{OS} regularizer is significantly more robust to different pruning rates. Even for pruning rates as high as $77\%$, the \emph{OS} method achieves a classification performance greater than $90\%$. We also investigate how the different models are affected at a pruning rate of $77\%$ using Grad-CAM visualizations. Figure \ref{fig:ablation_study} shows Grad-CAM activation maps for a ship image using the different methods. The AMC+\emph{OS} regularizer is still able to get better semantic localization of the object of interest despite the model being heavily pruned. This is not observed for the other baseline approaches.



\section{Conclusion}\label{sec:Conclusion}
In this work, we have studied a simple orthogonality constraint imposed on deep features, weakly motivated by studies in image formation physics. The proposed an orthogonal sphere ({\em OS}) regularization is not only much simpler than other complicated physical models but also produces promising benefits of practical interest such as diverse and robust representation. Furthermore, it can generate more plausible explanations in terms of semantically interpretable activation maps. In future work, it is promising to explore its combination with efficient deep-nets as recent efforts toward reducing computation and parameter storage costs.

\bibliography{bibfile1}
\end{document}